\RequirePackage[T1]{fontenc}
\documentclass[letterpaper,10pt,conference]{ieeeconf}
\usepackage{cite}
\usepackage{amsmath,amssymb,amsfonts}
\usepackage{graphicx}
\usepackage{textcomp}
\usepackage{xcolor}
\usepackage{booktabs}
\usepackage{array}
\usepackage{multirow}
\usepackage{microtype}
\usepackage{placeins}

\newcommand{\placeholderbox}[2]{%
  \fbox{\parbox[c][#1][c]{0.94\linewidth}{\centering\small\color{gray}#2}}}

\newcommand{\methodname}{FlowVLA}
\newcommand{\beliefvla}{\textsc{Belief-VLA}}

\begin{document}

\title{\LARGE \bf
\methodname{}: Learning Flow-Aware Physical Completion\\
for Quantitative Robotic Pouring
}

\author{%
Liwen Yang$^{1}$, Yutian Zhang$^{1}$, Jiaqi Zhai$^{2}$, Li Yang$^{3}$, Tutian Tang$^{3}$,\\
Qiwei Shan$^{1}$, Yancheng Bi$^{1}$, Qiankuan Shangguan$^{4}$, Yangkun Zhu$^{3}$,\\
Ce Hao$^{5}$, Dong Wei$^{2,\dagger}$, Qiaojun Yu$^{3,\dagger}$, Dibo Hou$^{1,\dagger}$\\[1mm]
{\small $^{1}$Zhejiang University \quad $^{2}$DEEP Robotics}\\
{\small $^{3}$Shanghai Artificial Intelligence Laboratory \quad $^{4}$Northeastern University, China}\\
{\small $^{5}$Beijing Zhongguancun Academy}\\
{\small $^{\dagger}$Corresponding authors: Dong Wei, Qiaojun Yu, and Dibo Hou}
}

\maketitle
\thispagestyle{empty}
\pagestyle{empty}

\begin{abstract}
During steady pouring, images and robot pose may remain nearly unchanged while transferred mass accumulates, making the return decision ambiguous for a whole-task vision--language--action (VLA) policy. We present FlowVLA, which separates motor generation from physical-completion estimation. A shared $\pi_{0.5}$ action expert learns approach, pour, and return with boundary-aware supervision. A target-independent estimator fuses visual, joint, wrist-wrench, and recurrent force-belief histories to maintain an estimate of transferred mass. The controller initiates return when this estimate plus predicted residual outflow reaches the requested mass. In coffee-pot/water experiments, FlowVLA achieves 13.1~g goal-balanced mean absolute error (MacroMAE), lower than all tested VLA baselines. Across all attempts, 53.7\% both complete the task and finish within $\pm10\%$ of the requested mass. The same model interpolates to held-out 150/250-g targets without retraining. Independently trained models further evaluate the formulation in three additional container--fluid settings.
\end{abstract}

\section{Introduction}
\label{sec:introduction}

\begin{figure}[t]
    \centering
    \includegraphics[width=\columnwidth]{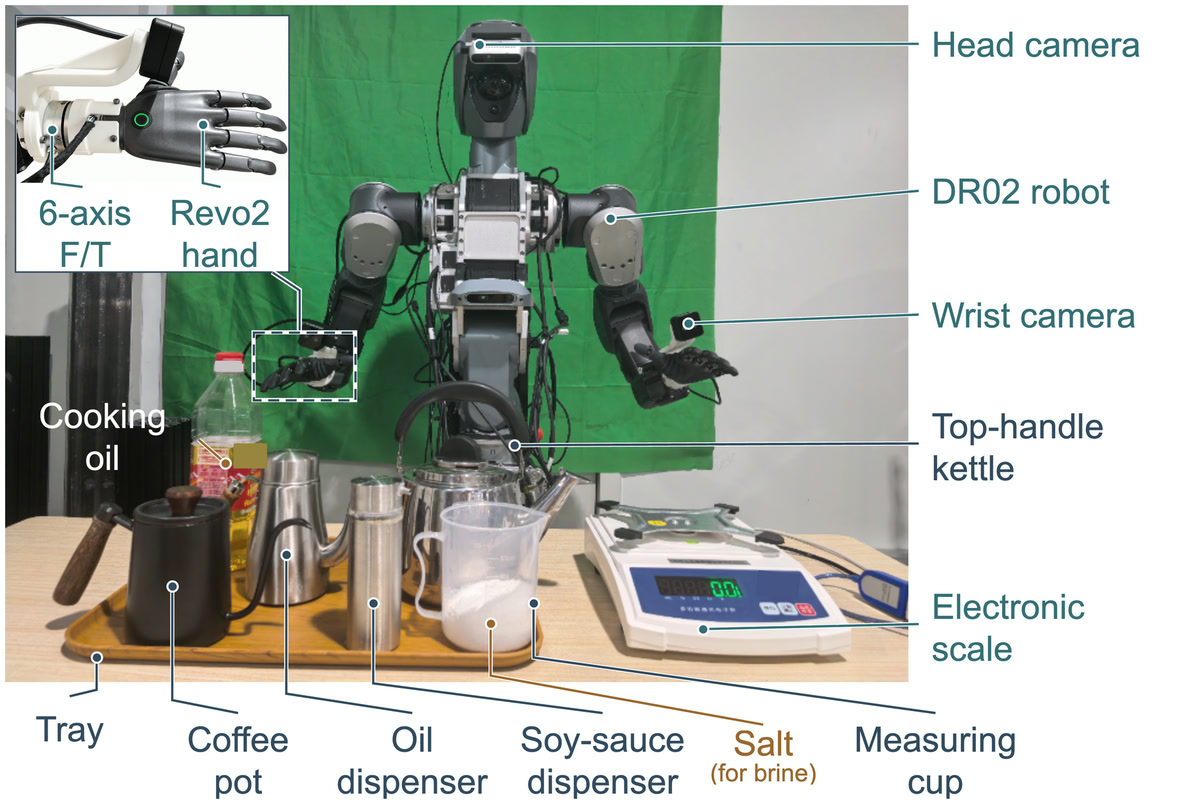}
    \centerfigcaptionsfalse
    \caption{Quantitative robotic pouring task. The robot performs grasping, pouring, return, and placement while targeting a requested transferred mass without online scale feedback.}
    \label{fig:setup_overview}
\end{figure}

Vision--language--action (VLA) models connect visual and language understanding to robot control, enabling a single policy to perform diverse manipulation tasks~\cite{zitkovich2023rt2,black2025pi05,bjorck2025groot}. Quantitative pouring adds a requirement beyond executing the manipulation sequence: the transferred mass must match a requested amount. This matters in food preparation~\cite{piacenza2022pouring} and household drink-serving tasks~\cite{zhu2023pour} (Fig.~\ref{fig:setup_overview}). Without online electronic-scale feedback, the robot must infer how much liquid has been transferred and when to return the container upright. During steady pouring, images and robot pose can remain nearly unchanged while mass accumulates, so visually similar states may require either continued pouring or return. We call this ambiguity \emph{physical-completion aliasing}. Understanding ``pour 200~g'' specifies the goal but does not reveal physical progress, and initiating return does not stop liquid transfer immediately.

\suppressfloats[t]

Whole-task VLA policies typically learn completion implicitly alongside motion. Policies use force or tactile feedback to guide action generation~\cite{yu2025forcevla,zhang2025tavla,xue2025rdp}, while FM-VLA encodes force histories as memory tokens~\cite{li2026fmvla}. Wrist wrench is informative about liquid transfer, but it also varies with arm motion, container pose, grasp geometry, and liquid dynamics~\cite{piacenza2022pouring}. Force conditioning therefore does not by itself define the accumulated physical quantity needed for a stopping decision. Precision-pouring controllers make this quantity explicit by closing the loop on fill level~\cite{schenck2017visual,zhu2023pour}, source-mass loss~\cite{piacenza2022pouring}, or predicted overpour~\cite{hu2025robocap}. Their feedback formulations offer a useful starting point, but generally assume prescribed grasps or constrained motions. The remaining challenge is to integrate quantitative completion into a learned whole-task policy that must also grasp, transport, pour, return, and place the container.

Our insight is to treat physical completion as a persistent state separate from motor generation. \methodname{} couples a shared $\pi_{0.5}$ action expert with explicit flow and mass estimation. Boundary-aware supervision distinguishes continued pouring from return at action-chunk boundaries. A recurrent force observer provides temporal belief to the action expert and a dedicated completion head, which fuses this belief with visual, joint, and wrist-wrench histories to estimate transferred mass. The physical estimators are target-independent, while the requested mass conditions the POUR prompt and completion threshold. The controller accounts for liquid transferred during return by triggering return when predicted terminal mass reaches the target. Real-robot comparisons with whole-task VLA and load-based baselines, mechanism ablations, and held-out target interpolation evaluate this formulation, alongside additional container--fluid regimes with independently trained models.

Our contributions are summarized as follows:
\begin{itemize}
    \item We formulate physical-completion aliasing in quantitative pouring and introduce \methodname{}, which separates motor generation from estimation of the hidden physical progress needed for quantitative stopping.
    \item We develop boundary-aware action learning that combines multirate force observations and recurrent belief with stage-specific prompts in a shared $\pi_{0.5}$ action expert. Row-wise masking and exact-return-boundary sampling jointly supervise the transition from steady pouring to return.
    \item We design a target-independent completion module that maintains flow and accumulated mass from causal multimodal histories and predicts residual outflow, enabling return decisions based on predicted terminal mass rather than current transferred mass.
\end{itemize}

\section{Related Work}
\label{sec:related_work}

\subsection{Robotic Precision Pouring}

Precision-pouring methods use visual feedback to estimate liquid level or volume in the receiving container and adjust the pouring trajectory~\cite{schenck2017visual,kennedy2019autonomous,zhu2023pour}. Other approaches plan with simplified fluid dynamics~\cite{pan2016motion} or acquire pouring skills through self-supervised trials~\cite{huang2021pouring}. These methods explicitly control the poured quantity, but generally assume a prescribed grasp or restricted pouring motions. Sensing at the source container offers an alternative to observing the liquid surface. Pouring by Feel estimates mass loss from proprioceptive and tactile measurements and regulates pouring with proportional--integral--derivative (PID) control~\cite{piacenza2022pouring}. RoboCAP is the closest quantity-aware reference: it accumulates capacitance-based mass estimates and predicts overpour to stop a prescribed wrist motion~\cite{hu2025robocap}. Flow estimation and delayed-outflow compensation also appear in industrial ladle control~\cite{noda2007pouring,sueki2021flow}. Like these methods, \methodname{} reasons in physical units. Unlike prescribed-motion controllers, it couples completion estimation to a learned sequence spanning grasping, transport, pouring, return, and placement.

\subsection{Force-Aware and Progress-Aware Robot Policies}

Diffusion Policy learns visually conditioned action sequences through denoising diffusion~\cite{chi2023diffusionpolicy}. VLA models also condition actions on language. RT-2 transfers vision--language knowledge to robotic control~\cite{zitkovich2023rt2}, while $\pi_{0.5}$ and GR00T N1 learn language-conditioned manipulation across tasks~\cite{black2025pi05,bjorck2025groot}. Force-aware methods add feedback about physical interaction. Reactive Diffusion Policy (RDP) combines slower visual action generation with high-rate tactile correction~\cite{xue2025rdp}. Within VLA models, ForceVLA incorporates six-axis wrench through a mixture of experts~\cite{yu2025forcevla}, and TA-VLA learns from torque histories with auxiliary future-torque prediction~\cite{zhang2025tavla}. ForceVLA2 further couples force awareness with hybrid force--position control~\cite{li2026forcevla2}. FM-VLA is the closest temporal-force reference: it encodes force histories as memory tokens for action generation~\cite{li2026fmvla}. \methodname{} instead converts causal multimodal history into explicit flow and mass estimates and uses them in a terminal-mass rule, separating physical completion from action generation. Beyond physical feedback, progress-aware policies estimate the remaining task horizon or learn visual progress representations~\cite{kim2026progvla,yan2026progressvla}. These help locate the robot within an execution sequence, but normalized progress does not specify how many grams have been transferred. \methodname{} maintains flow in g/s and mass in g across perceptually similar states, then estimates whether returning now will achieve the requested quantity. For persistent state estimation, differentiable filters and KalmanNet combine learned observations or updates with explicit state evolution~\cite{lee2020multimodal,revach2022kalmannet}. Our observer integrates flow and recursively updates mass and uncertainty.

\section{Method}
\label{sec:method}
\begin{figure*}[t]
    \centering
    \IfFileExists{figures/submission/fig1_pour_vla_architecture.jpg}{%
        \includegraphics[width=0.98\textwidth]{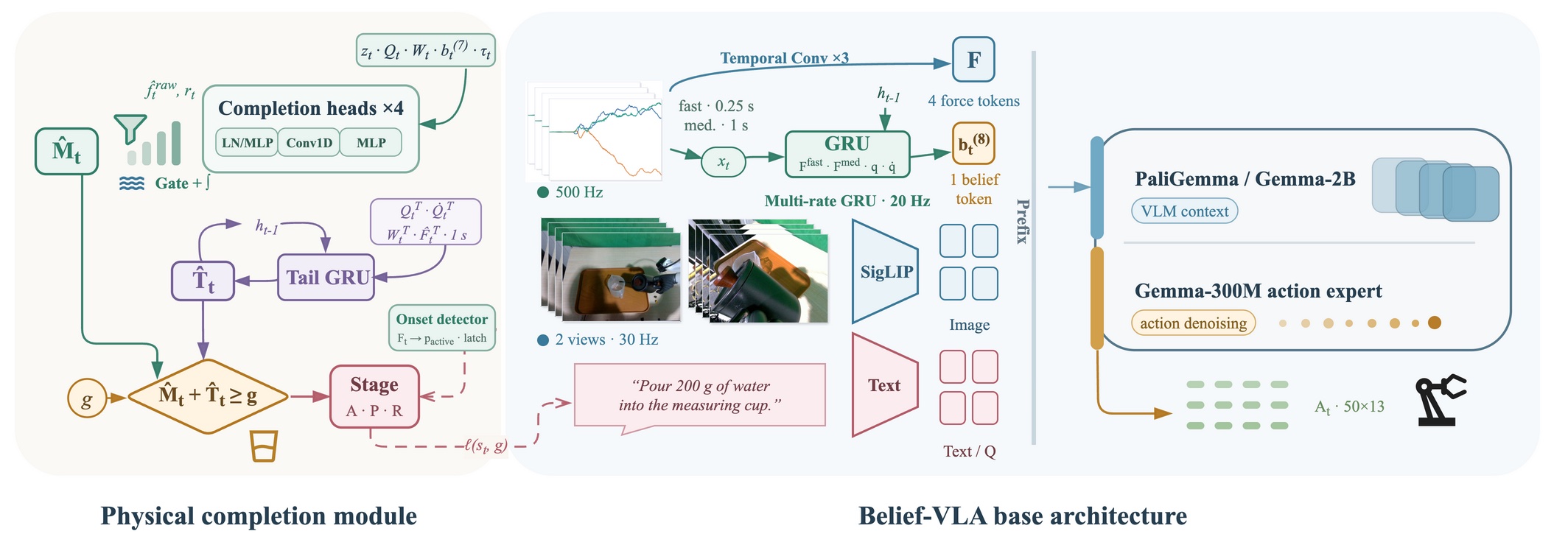}
    }{%
        \placeholderbox{1.48in}{Figure 1 placeholder: two RGB views, 13-D state, a 128-by-3 base-frame force window, and an eight-dimensional recurrent belief condition a boundary-aware $\pi_{0.5}$ action expert. The physical path estimates monotonic progress and a scalar return tail from causal visual, joint, wrench, belief, and frozen-flow histories. The controller compares their sum with the target and selects APPROACH, POUR, or RETURN.}
    }
    \caption{The architecture of \methodname{}. The action backbone fuses 30-Hz images and robot state with 500-Hz force history and a 20-Hz recurrent belief. The physical-completion module estimates transferred mass and return tail, whose sum determines when to return.}
    \label{fig:architecture}
\end{figure*}

\subsection{Problem Formulation}
\label{sec:formulation}

Given target mass $g\in\mathbb{R}_{+}$, the robot must grasp a source container, pour into a receiver, return upright, and place the source down.

At policy step $t$, $I_t=(I_t^{\mathrm{head}},I_t^{\mathrm{wrist}})$ denotes two RGB views and $q_t\in\mathbb{R}^{13}$ the active right-arm and hand joints. The action expert receives these inputs, a causal base-frame force window $F_t^{\mathrm{base}}$, and belief $b_t^{(8)}$ from a frozen recurrent observer. It predicts absolute joint-position chunks $A_t\in\mathbb{R}^{50\times13}$ at 30~Hz.

Let $M_t$ and $f_t$ denote cumulative transferred mass and instantaneous mass flow. Neither is directly observed online. During steady pouring, two states can satisfy
\begin{equation}
 (I_{t_1},q_{t_1})\approx(I_{t_2},q_{t_2}),\quad
 M_{t_1}\neq M_{t_2},\quad a_{t_1}^{*}\neq a_{t_2}^{*}.
 \label{eq:aliasing}
\end{equation}
The relation in \eqref{eq:aliasing} describes \emph{physical-completion aliasing}: similar observations can require continued pouring or return.

The stage $s_t\in\{\mathsf{A},\mathsf{P},\mathsf{R}\}$ denotes approach, pour, or return. The learned branches are
\begin{equation}
\begin{aligned}
 b_t^{(8)} &= B_{\omega}(F_{\leq t}^{\mathrm{base}},q_{\leq t}),\\
 A_t &\sim \pi_{\theta}(I_t,q_t,F_t^{\mathrm{base}},b_t^{(8)},\ell(s_t,g)),\\
 (\widehat f_t^{\mathrm{raw}},r_t) &= R_{\phi}(z_t,Q_t,W_t,b_t^{(7)},\tau_t),\\
 \widehat T_t &= H_{\eta}(Q_t^{T},\dot Q_t^{T},W_t^{T},\widehat F_t^{T}).
\end{aligned}
\label{eq:factorization}
\end{equation}
In \eqref{eq:factorization}, $z_t$, $Q_t$, and $W_t$ denote visual features, joint history, and sensor-frame wrench history, while $b_t^{(7)}$ omits observer reliability and $\tau_t$ is time since predicted onset. Superscript $T$ identifies tail-prediction histories (Sec.~\ref{sec:completion}). The goal enters only the POUR prompt and comparator, never $B_{\omega}$, $R_{\phi}$, or $H_{\eta}$.

For a demonstrated standard return from annotated onset $t_3$ to complete flow cutoff $t_4$, the tail target is
\begin{equation}
 T=M(t_4)-M(t_3)=\int_{t_3}^{t_4}f(t)\,dt.
 \label{eq:tail_target}
\end{equation}

These relations define three design principles. Persistent physical state carries progress through visually similar observations. Boundary-aware action learning separates the motor modes needed to continue pouring and initiate return. Predictive completion accounts for liquid that will still leave the source during return. Sections~\ref{sec:state_action} and~\ref{sec:completion} describe how we implement these principles.

\subsection{Persistent State and Boundary-Aware Action Generation}
\label{sec:state_action}

\textbf{Persistent physical state.}
A current observation does not reveal how much liquid has already been transferred, so the policy needs a state that persists over time (Fig.~\ref{fig:architecture}). The recurrent observer supplies this belief at 20~Hz. A single wrist-force sample cannot reliably separate liquid flow from arm inertia and robot vibration. The observer therefore combines a fast branch, which encodes 32 force samples over 0.25~s and their derivatives, with a medium branch over 32 low-pass samples spanning 1~s. A gated recurrent unit (GRU)~\cite{cho2014learning} fuses these signals with 13-D joint position and finite-difference velocity and carries information across stages. The resulting belief is
\begin{equation}
 b_t^{(8)}=[\widehat f_t^{\mathrm{bel}},\sigma_t^f,\widehat M_t^{\mathrm{bel}},\sigma_t^M,\beta_t^f,\rho_t,p_t^{\mathrm{active}},l_t^{\mathrm{onset}}].
 \label{eq:belief_vector}
\end{equation}
The entries in \eqref{eq:belief_vector} describe three aspects of the physical state. Progress is recursively integrated mass $\widehat M_t^{\mathrm{bel}}$. Flow state comprises corrected flow $\widehat f_t^{\mathrm{bel}}$, slowly updated bias $\beta_t^f$, flow-active probability $p_t^{\mathrm{active}}$, and onset latch $l_t^{\mathrm{onset}}$. Estimate quality comprises predicted flow standard deviation $\sigma_t^f$, propagated mass standard deviation $\sigma_t^M$, and exponentially smoothed force reliability $\rho_t$. Superscript $\mathrm{bel}$ distinguishes observer estimates from completion-head outputs. The latch activates after three consecutive valid updates with activity-weighted flow $\geq0.3$~g/s and active probability $\geq0.3$, and persists until episode reset. The observer is independently trained on teleoperated demonstrations with offline flow supervision, then frozen. In \beliefvla{}, the latch is a feature, not a stage-transition trigger.

\textbf{Boundary-aware action learning.}
The policy must distinguish continued pouring from return rather than blend these motor modes within an action chunk. Inspired by $\pi_{0.5}$'s multimodal action generation~\cite{black2025pi05}, we extend its pretrained action expert with four force tokens and one belief token in the image and text prefix. Temporal convolutions encode the causal $128\times3$ base-frame force window, and a multilayer perceptron (MLP) maps the eight-dimensional belief into one token. These tokens, the stage prompt, and discretized joint state condition 50-step arm--hand action chunks.

Prompts specify the container--fluid domain and current stage. Only the POUR prompt includes the target mass. Let $d$ denote the domain and $\mathcal G_d$ its training targets:
\begin{equation}
 \mathcal{L}_d=
 \{\ell_{\mathsf A}(d),\ell_{\mathsf R}(d)\}
 \cup
 \{\ell_{\mathsf P}(d,g)\mid g\in\mathcal G_d\}.
 \label{eq:prompt_set}
\end{equation}
The three stage templates in \eqref{eq:prompt_set} yield $|\mathcal L_d|=|\mathcal G_d|+2$ prompt instances. In the primary coffee-pot/water domain, $\mathcal G_d=\{100,200,300\}\,\mathrm{g}$ gives five strings.

A standard 50-step action target may cross a stage boundary and contain incompatible motion modes. Our boundary-aware supervision combines row-wise masking with exact-return-boundary sampling. We retain each anchor but mask action rows after the next stage boundary. With valid-row mask $m_{t,h}\in\{0,1\}$, the flow-matching loss is
\begin{equation}
 \mathcal{L}_{\mathrm{act}}=
 \frac{\sum_{h=0}^{H-1}m_{t,h}\lVert v_{\theta,t,h}-u_{t,h}\rVert_2^2}
 {32\sum_{h=0}^{H-1}m_{t,h}},\qquad H=50.
 \label{eq:masked_action_loss}
\end{equation}
The loss in \eqref{eq:masked_action_loss} supervises all 32 internal coordinates: 13 normalized joint targets and 19 zero-padded coordinates. Deployment denormalizes and executes only the 13 active coordinates.
An exact boundary anchor is the first desired return frame and retains the complete return horizon.

\subsection{Structured Physical-Completion Estimation}
\label{sec:completion}

\textbf{Predictive completion.}
Completion depends on both the mass already transferred and the residual mass that will leave the source during return. The physical-completion module combines a completion head, a hard-zero gate, temporal integration, and a tail predictor to estimate these quantities and compare their sum with the target.

The completion head reuses visual features $z_t$, pooled to $2\times2$ from the frozen image encoder's tokens before stage-text conditioning. Its other inputs are joint history $Q_t\in\mathbb{R}^{8\times13}$, sensor-frame wrench history $W_t\in\mathbb{R}^{512\times6}$, eight steps of $b_t^{(7)}$, and elapsed time $\tau_t$. This six-axis wrench path differs from the action model's shorter three-axis base-frame force window.

Projected visual features, temporally encoded wrench, and joint/belief summaries feed a fusion network that predicts flow (g/s) and a mass correction (g):
\begin{equation}
 \widehat f_t^{\mathrm{raw}}=20\,\mathrm{softplus}(z_t^f),\qquad
 r_t=8\,\tanh(z_t^r).
 \label{eq:router_outputs}
\end{equation}
In \eqref{eq:router_outputs}, $z_t^f$ and $z_t^r$ are the fusion network's scalar outputs.

Integration makes even small false-flow predictions before onset or after cutoff accumulate into mass error. A hard-zero gate suppresses this drift. Four temporal gates use joint, belief, and wrench histories, sharing seeds with the completion-head ensemble. Each maps logit $a_t^j$ to openness $o_t^j=\operatorname{sigmoid}(-a_t^j)$. The gate in \eqref{eq:hard_gate} retains flow only when mean openness reaches 0.5:
\begin{equation}
 \begin{aligned}
 \widehat f_t&=\mathbf{1}\!\left[\tfrac14\sum_j o_t^j\geq0.5\right]
 \left(\tfrac14\sum_j\widehat f_{t,j}^{\mathrm{raw}}\right),\\
 r_t&=\tfrac14\sum_j r_{t,j}.
 \end{aligned}
 \label{eq:hard_gate}
\end{equation}

During APPROACH, we buffer the latest four seconds of predicted flow. When onset is latched and POUR begins, the buffered flow integral plus $r_t$ initializes mass, accounting for transfer before onset detection. The gated flow estimate $\widehat f_t$ and correction $r_t$ update the integral and monotonic mass estimate in \eqref{eq:conservative_progress}:
\begin{equation}
 \begin{aligned}
 \widetilde M_t&=\widetilde M_{t-1}+
 \tfrac12(\widehat f_{t-1}+\widehat f_t)\Delta t,\\
 \widehat M_t&=\max(\widehat M_{t-1},\widetilde M_t+r_t,0).
 \end{aligned}
 \label{eq:conservative_progress}
\end{equation}

Returning the container does not stop liquid transfer immediately. A causal GRU predicts the resulting residual mass $\widehat T_t$. It encodes 31 samples over one second of joint position, velocity, six-axis wrench, and the frozen completion head's ungated ensemble-mean flow ($\widehat F_t^T$ in \eqref{eq:factorization}). An MLP with Softplus outputs a nonnegative scalar in grams. Return begins when
\begin{equation}
 \widehat M_t+\widehat T_t\geq g.
 \label{eq:return_trigger}
\end{equation}
The criterion in \eqref{eq:return_trigger} triggers return based on predicted terminal mass rather than current transferred mass.
Before a full second of history is available, the controller uses $\widehat M_t\geq g$. If flow remains below 0.5~g/s and progress increases by less than 0.25~g for 3~s, it requests one same-stage retry. A second stall triggers a safe return and counts as a failure.

\subsection{Supervision and Execution}
\label{sec:physical_supervision}

The belief observer is trained with Huber losses on flow and mass, multi-window integral and signed-bias penalties, and a Gaussian negative log-likelihood (NLL) for flow uncertainty. Binary cross-entropy (BCE) supervises activity, with positive labels during the annotated flowing interval and negative labels before onset and after cutoff. Reliability and mass-observation uncertainty are learned through the recurrent mass update without separate labels.

The completion-head objective combines confidence-weighted flow and mass interval losses, false-flow suppression, long-horizon positive-bias penalties, and residual regularization. Interval losses use $d(x,[l,u])=\max(l-x,0)+\max(x-u,0)$; mass losses use trapezoidal integration in physical units. The primary ensemble averages predictions from four separately trained completion heads and remains frozen during automatic evaluation.

Gate training uses confidence-weighted BCE: valid pre-onset labels with $|f|\leq0.25$~g/s supervise closure, and valid labels with $f>1$~g/s supervise opening. Other frames are excluded from classification. Auxiliary losses penalize zero-flow leakage and its accumulated mass, protect nonzero flow, and preserve the frozen completion head's flow and mass predictions. A straight-through estimator supplies gradients through the hard decision.

Tail supervision is extracted once per episode at the annotated return boundary using \eqref{eq:tail_target}. Training uses a 1-g Huber loss, and validation MacroMAE selects the checkpoint.

During execution, the controller advances from APPROACH to POUR and then to RETURN, where it remains until the episode ends. The observer's onset latch triggers POUR; \eqref{eq:return_trigger} or the stall branch triggers RETURN. Each transition selects a new prompt, invalidates the previous action chunk, and requests a fresh prediction.

\section{Experiments}
\label{sec:experiments}
\begin{figure*}[t]
    \centering
    \begin{minipage}[c]{0.755\textwidth}
        \centering
        \includegraphics[width=\linewidth]{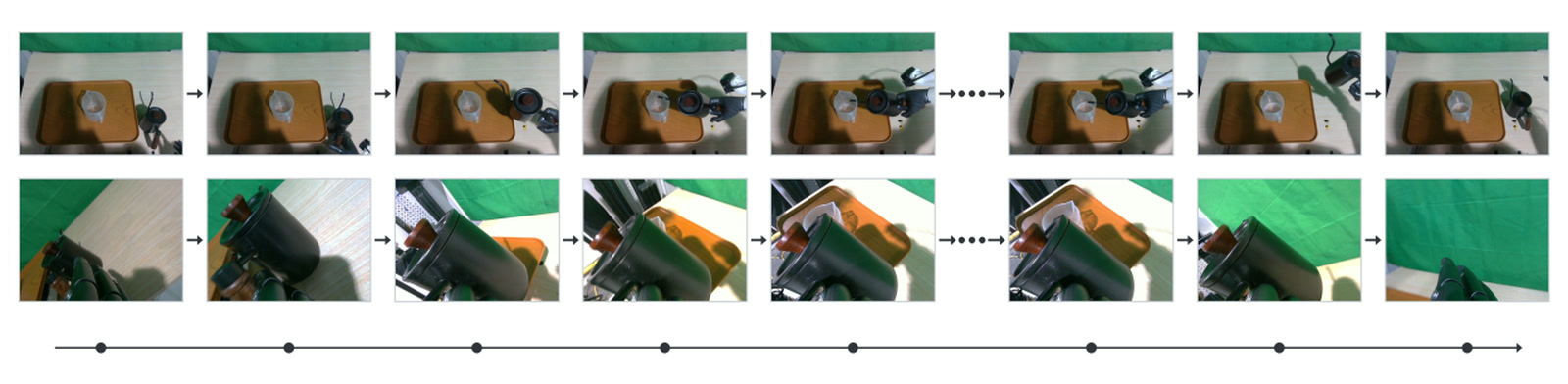}
    \end{minipage}\hfill
    \begin{minipage}[c]{0.225\textwidth}
        \centering
        {\scriptsize (a)}\includegraphics[width=0.38\linewidth,height=0.55in,keepaspectratio]{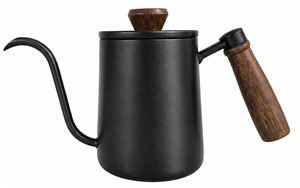}\hfill
        {\scriptsize (b)}\includegraphics[width=0.38\linewidth,height=0.55in,keepaspectratio]{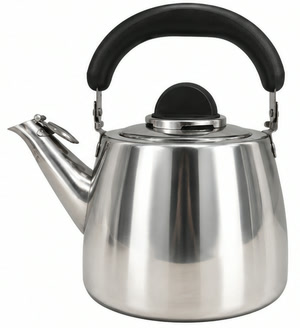}\\[-0.5mm]
        {\scriptsize (c)}\includegraphics[width=0.38\linewidth,height=0.55in,keepaspectratio]{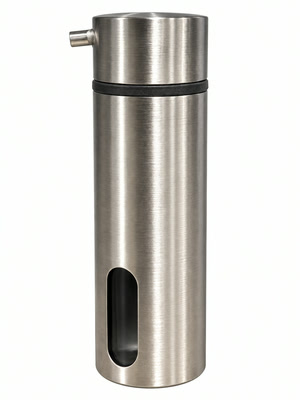}\hfill
        {\scriptsize (d)}\includegraphics[width=0.38\linewidth,height=0.55in,keepaspectratio]{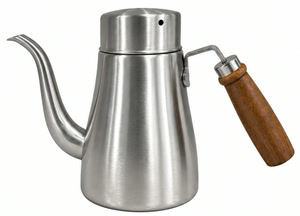}
    \end{minipage}
    \caption{Left: views from head and wrist cameras during approach, grasp, steady pouring, return, and placement. Right: (a) coffee pot, (b) top-handle kettle, (c) soy-sauce dispenser, and (d) oil dispenser.}
    \label{fig:task_sequence}
\end{figure*}

\begin{figure*}[t]
    \centering
    \includegraphics[width=0.91\textwidth]{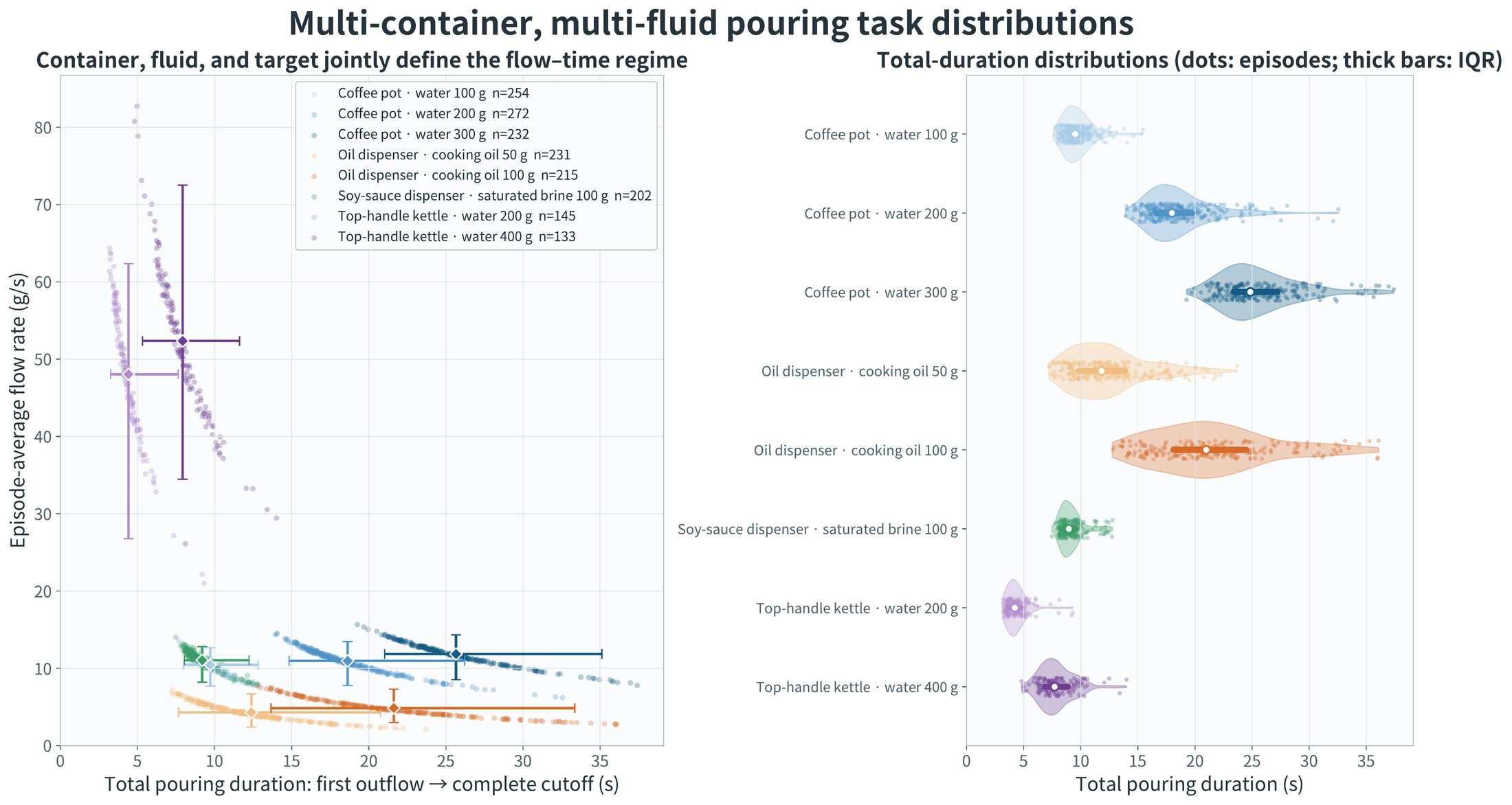}
    \caption{Flow-labeled demonstration datasets for the four domains (1,684 episodes total). Left: episode mean flow versus duration; diamonds are task means and error bars span the central 95\% episode ranges. Right: duration distributions; thick lines mark interquartile ranges.}
    \label{fig:training_flow_distribution}
\end{figure*}

\subsection{Experimental Setup}
\label{sec:experimental_setup}

\textbf{Platform and tasks.}
Experiments use a DR02 upper-body robot with Revo2 hands. The single-arm policy controls seven arm joints and six active hand coordinates. Observations comprise 30-Hz head and wrist images and 500-Hz wrist wrench from a six-axis wrist force/torque sensor. Demonstrations are collected by teleoperation. A 10-Hz electronic scale beneath the receiving container, with 0.1-g specified accuracy, supplies offline supervision and final evaluation but is unavailable to the deployed policy.

The primary comparison uses a coffee pot with water at 100/200/300~g. Targets of 150/250~g are withheld from training and tuning. Additional domains use a top-handle kettle with water, a soy-sauce dispenser with saturated brine, and an oil dispenser with cooking oil. Saturated brine is prepared near $26\,^{\circ}$C and may retain slight salt sediment. Figure~\ref{fig:task_sequence} shows the task and source containers.

\textbf{Data and baselines.}
The learned-policy comparison uses a common 758-episode coffee-pot/water dataset: 254, 272, and 232 episodes at 100, 200, and 300~g, respectively. The additional datasets contain 278 demonstrations for top-handle kettle/water (145/133 at 200/400~g), 202 for soy-sauce dispenser/saturated brine at 100~g, and 446 for oil dispenser/cooking oil (231/215 at 50/100~g). We train an independent \methodname{} model for each additional domain using the same target-stratified splitting rule.

Each container--fluid domain uses its own teleoperated dataset to train the physical estimators. Within each target, validation and test sets each contain $\lfloor0.1N\rfloor$ episodes, with the remainder used for training. The primary domain yields a fixed 608/75/75 split shared by the completion head and tail predictor. Normalization uses training data only, and model selection uses the validation partition. Flow and mass labels are reconstructed offline from electronic-scale data.

Primary-domain comparisons include $\pi_{0.5}$~\cite{black2025pi05}, its matched-compute data-scale control, ForceVLA~\cite{yu2025forcevla}, TA-VLA-Wrench, and \beliefvla{}. Only the $\pi_{0.5}$ data-scale control uses the additional 695 coffee-pot/water demonstrations, for 1,453 episodes in total. TA-VLA-Wrench follows TA-VLA's torque-history interface and future-effort objective, using ten causal wrist-wrench samples on our platform~\cite{zhang2025tavla}. \beliefvla{} is our independently trained whole-task variant without stage routing, boundary masking, or the explicit completion rule. \beliefvla{} and \methodname{} each receive 20,000 action updates with global batch 32.

\begin{table*}[t]
\centering
\caption{Primary coffee-pot/water results at 100/200/300~g. TR-LT replays held-out paths in their original layouts; learned policies use randomized initial source positions (Sec.~\ref{sec:experimental_setup}).}
\label{tab:automatic_results}
\footnotesize
\setlength{\tabcolsep}{1.5pt}
\begin{tabular*}{0.98\textwidth}{@{\extracolsep{\fill}}lrrrrllrrrr@{}}
\toprule
\multirow{2}{*}{Method} & \multirow{2}{*}{Grasp} & \multirow{2}{*}{\shortstack{Post-grasp\\success}} & \multirow{2}{*}{E2E} & \multirow{2}{*}{$N_m$} & \multirow{2}{*}{\shortstack{$N_{\mathrm{mass}}$\\(100/200/300)}} & Per-goal MAE (g) & \multirow{2}{*}{Pooled} & \multirow{2}{*}{Macro} & \multirow{2}{*}{$\leq10\%$} & \multirow{2}{*}{E2E+$\pm10\%$} \\
\cmidrule(lr){7-7}
 & & & & & & 100/200/300 & & & & \\
\midrule
\multicolumn{11}{@{}l}{\textit{Trajectory-replay controls}}\\
TR-LT-ID & 100.0\% & 100.0\% & 100.0\% & 37 & 37 (12/15/10) & 66.86/95.48/124.33 & 94.0 & 95.6 & 5.4\% & 5.4\% \\
TR-LT-Cal & 100.0\% & 100.0\% & 100.0\% & 38 & 38 (13/12/13) & 13.40/21.17/25.04 & 19.8 & 19.9 & 57.9\% & 57.9\% \\
\midrule
\multicolumn{11}{@{}l}{\textit{External baselines}}\\
$\pi_{0.5}$ & 65.2\% & 33.3\% & 21.7\% & 45 & 15 (7/6/2) & 69.34/54.97/75.40 & 64.4 & 66.6 & 13.3\% & 2.9\% \\
$\pi_{0.5}$, 1,453 demos & 73.3\% & 36.4\% & 26.7\% & 33 & 12 (6/5/1) & 72.06/66.32/80.20 & 70.3 & 72.9 & 8.3\% & 2.2\% \\
ForceVLA & 64.3\% & 70.4\% & 45.2\% & 54 & 38 (12/10/16) & 34.38/56.65/59.83 & 51.0 & 50.3 & 23.7\% & 10.7\% \\
TA-VLA-Wrench & 63.6\% & 73.8\% & 47.0\% & 42 & 31 (11/13/7) & 141.05/109.08/54.51 & 108.1 & 101.6 & 12.9\% & 6.1\% \\
\midrule
\multicolumn{11}{@{}l}{\textit{Our variants}}\\
\beliefvla{} (whole-task) & 67.0\% & 88.1\% & 59.0\% & 67 & 59 (16/17/26) & 14.87/22.47/42.25 & 29.1 & 26.5 & 52.5\% & 31.0\% \\
\methodname{} & 68.3\% & 94.0\% & 64.2\% & 84 & 79 (28/26/25) & 6.96/13.81/18.66 & 12.9 & 13.1 & 83.5\% & 53.7\% \\
\bottomrule
\end{tabular*}
\par\vspace{1mm}\noindent
\parbox{0.98\textwidth}{\scriptsize $N_m$ counts successful grasps; $N_{\mathrm{mass}}$ counts completed trials with final mass. The $\leq10\%$ rate uses $N_{\mathrm{mass}}$; E2E+$\pm10\%$ counts completed trials within $\pm10\%$ of the target over all attempts. MacroMAE weights goals equally.}
\end{table*}

\begin{figure*}[!t]
    \centering
    \includegraphics[width=0.98\textwidth]{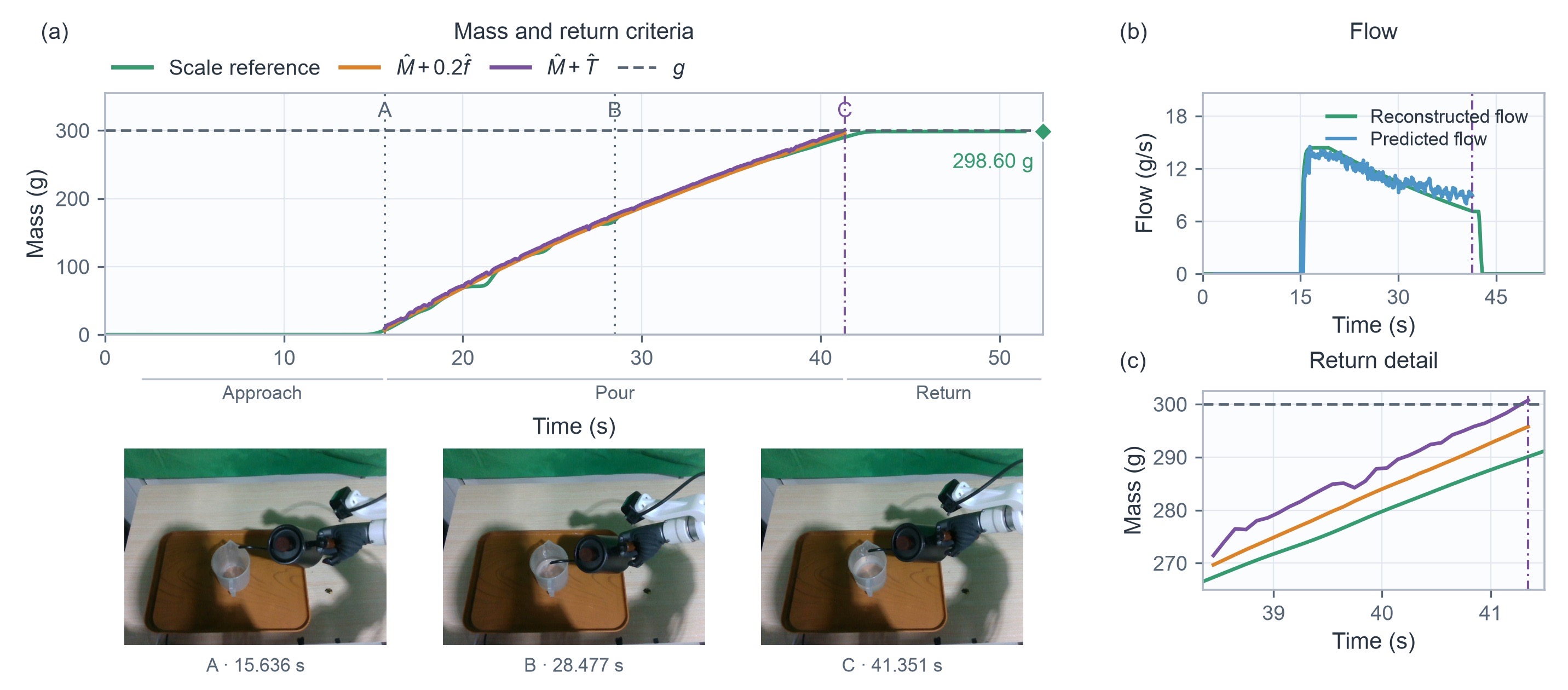}
    \caption{Automatic 300-g trial. (a) Predicted terminal mass using fixed and learned tails, offline electronic-scale reference, and frames at onset, mid-pour, and return. (b) Reconstructed flow and completion-head prediction. (c) Return decision.}
    \label{fig:closed_loop_300g}
\end{figure*}

Two \emph{Trajectory Replay + Load Threshold} (TR-LT) controls replay held-out teleoperated paths in the original layouts of the source and receiving containers. They replay the recorded approach and return/placement segments, defined by joint and action changes, but replace the demonstrated POUR duration: after approach, the source holds its pouring pose until an online load threshold triggers return. VLA trials instead randomize the initial source-container position within a bounded region.

TR-LT uses the same target-stratified 608/75/75 split. The 75 test paths (25/27/23 at 100/200/300~g) are randomly assigned to disjoint groups: 37 for TR-LT-ID and 38 for TR-LT-Cal. After the source reaches steady outflow, $\mathbf F_{\mathrm{ref}}$ is the median base-frame force over 0.25--0.75~s, $\mathbf u=\mathbf F_{\mathrm{ref}}/\|\mathbf F_{\mathrm{ref}}\|$, and $\mathbf F_t^{\mathrm{med}}$ is the median over the 0.5-s window ending 0.25~s before $t$. Return is triggered by
\begin{equation}
 x_t=\frac{1000}{g_0}(\mathbf F_{\mathrm{ref}}-\mathbf F_t^{\mathrm{med}})^\top\mathbf u,
 \qquad ax_t+b+c_{\mathrm{tail}}\geq g,
 \label{eq:mp_load_trigger}
\end{equation}
In \eqref{eq:mp_load_trigger}, $g_0$ is gravitational acceleration and $x_t$ is the mass-equivalent axial load decrease from the steady-pour reference. \emph{TR-LT-ID} uses the identity mapping, $a=1,b=0$. \emph{TR-LT-Cal} fits $x_t$ to electronic-scale cumulative mass by Huber regression on the 608 training demonstrations, yielding $a=1.26319$ and $b=33.69355$~g. Both use the training median return-to-cutoff mass, $c_{\mathrm{tail}}=5.38932$~g. These target-independent parameters are frozen before testing.

\begin{table*}[t]
\centering
\caption{System-level mechanism ablations: mass accuracy and manipulation success.}
\label{tab:ablations}
\footnotesize
\setlength{\tabcolsep}{3pt}
\resizebox{0.98\textwidth}{!}{%
\begin{tabular}{@{}llrrllrrrrrr@{}}
\toprule
Component & Variant & $N_a$ & $N_m$ & \shortstack{$N_{\mathrm{mass}}$\\100/200/300} & \shortstack{Per-goal MAE (g)\\100/200/300} & Pooled & Bias & Macro & Grasp & \shortstack{Post-grasp\\success} & E2E \\
\midrule
Full & $\widehat M_t+\widehat T_t\geq g$ & 123 & 84 & 79 (28/26/25) & 6.96/13.81/18.66 & 12.92 & $+2.65$ & 13.15 & 68.3\% & 94.0\% & 64.2\% \\
\midrule
Action & No boundary-aware supervision & 59 & 39 & 33 (13/11/9) & 13.39/18.57/28.84 & 19.33 & $+12.16$ & 20.27 & 66.1\% & 84.6\% & 55.9\% \\
State & $-$hard-zero gate & 53 & 37 & 35 (12/12/11) & 12.47/13.46/24.82 & 16.69 & $-5.73$ & 16.92 & 69.8\% & 94.6\% & 66.0\% \\
Trigger & $\widehat M_t\geq g$ & 61 & 40 & 39 (13/12/14) & 10.18/20.11/26.89 & 19.24 & $+6.93$ & 19.06 & 65.6\% & 97.5\% & 63.9\% \\
 & $\widehat M_t+0.2\widehat f_t\geq g$ & 59 & 40 & 38 (12/13/13) & 7.26/13.86/26.24 & 16.01 & $+2.74$ & 15.79 & 67.8\% & 95.0\% & 64.4\% \\
\bottomrule
\end{tabular}}
\par\vspace{1mm}\noindent
\parbox{0.98\textwidth}{\scriptsize $N_a$: attempts; $N_m$: successful grasps. Mass errors are in grams and use completed trials with recorded final mass. MacroMAE weights goals equally. Success-rate definitions follow Table~\ref{tab:automatic_results}.}
\end{table*}

\begin{table*}[t]
\centering
\caption{\methodname{} on held-out targets and additional container--fluid tasks. Held-out targets use the unchanged primary model; additional tasks use independently trained domain-specific models.}
\label{tab:domain_results}
\footnotesize
\setlength{\tabcolsep}{3pt}
\resizebox{0.98\textwidth}{!}{%
\begin{tabular}{@{}lrrrrrrrrrr@{}}
\toprule
Container / fluid & Goals (g) & $N_a$ & $N_m$ & $N_{\mathrm{mass}}$ & MAE (g) & Macro Rel. & Grasp & Post-grasp & E2E & E2E+$\pm10\%$ \\
 & & & & & & MAE & & success & & \\
\midrule
Coffee pot / water & 150/250 & 31 & 22 & 10/10 & 10.94/19.71 & 7.6\% & 71.0\% & 90.9\% & 64.5\% & 54.8\% \\
Top-handle kettle / water & 200/400 & 32 & 25 & 12/12 & 34.54/50.73 & 15.0\% & 78.1\% & 96.0\% & 75.0\% & 25.0\% \\
\shortstack[l]{Soy-sauce dispenser /\\saturated brine} & 100 & 31 & 18 & 11 & 9.05 & 9.1\% & 58.1\% & 61.1\% & 35.5\% & 25.8\% \\
Oil dispenser / cooking oil & 50/100 & 47 & 30 & 14/13 & 4.23/11.86 & 10.2\% & 63.8\% & 90.0\% & 57.4\% & 31.9\% \\
\bottomrule
\end{tabular}}
\par\vspace{1mm}\noindent
\parbox{0.98\textwidth}{\scriptsize $N_a$: all attempts; $N_m$: successful grasps. $N_{\mathrm{mass}}$ and MAE are listed in goal order. Macro Rel. MAE averages $\mathrm{MAE}_g/g$ equally across goals. Success-rate definitions follow Table~\ref{tab:automatic_results}.}
\end{table*}

\textbf{Protocol and metrics.}
All methods use the same success criteria, with randomized method and goal order. After grasping and flow onset, the controller must initiate return within 50~s, avoid excessive spillage and tipping of the receiver, tray, or source, and place and release the source stably. Late or incorrect return, collision, a safety stop, a second stall, or human override is an end-to-end (E2E) failure. We report manipulation completion and quantitative accuracy separately. Overall quantitative success combines both requirements over all attempts. Let $N_a$, $N_m$, and $N_{\mathrm{E2E}}$ count attempts, successful grasps, and completed tasks. Grasp, post-grasp, and E2E success are $N_m/N_a$, $N_{\mathrm{E2E}}/N_m$, and $N_{\mathrm{E2E}}/N_a$, respectively.

\pagebreak[4]
For completed trials with recorded final mass ($N_{\mathrm{mass}}$), we report pooled mean absolute error (MAE), computed by averaging $|M_{\mathrm{final}}-g|$ over trials, goal-balanced $\mathrm{MacroMAE}=(\mathrm{MAE}_{100}+\mathrm{MAE}_{200}+\mathrm{MAE}_{300})/3$, signed bias and the fraction within $\pm10\%$. For overall quantitative success, accuracy requires final mass within $\pm10\%$ of the target.

Flow MAE for the gated completion-head output is measured against reconstructed references at matched timestamps before recorded return onset, using valid flow labels. Errors are time-weighted within episodes and averaged equally across episodes.

We compare explicit completion with whole-task policies and data scaling, test calibrated load sensing under prescribed motion, ablate the three core mechanisms, and evaluate held-out targets and additional flow regimes.

\subsection{Main Results and Mechanism Analysis}

Force feedback primarily improves whether the policy terminates pouring, but not necessarily whether it knows when the requested quantity has been reached. Compared with $\pi_{0.5}$, ForceVLA and TA-VLA-Wrench increase post-grasp success from 33.3\% to 70.4\% and 73.8\%, respectively, but their MacroMAEs remain 50.3 and 101.6~g. \beliefvla{} further provides the action expert with a recurrent physical belief containing flow and mass information, reaching 88.1\% post-grasp success and 26.5~g MacroMAE. This suggests that recurrent belief improves the policy's representation of pouring progress. With explicit completion estimation and triggering, \methodname{} further reduces MacroMAE to 13.1~g. Overall quantitative success accordingly rises from 31.0\% for \beliefvla{} to 53.7\% for \methodname{} (Table~\ref{tab:automatic_results}).

\begin{samepage}
With zero-flow gating, the completion head achieves a flow MAE of 0.38~g/s on the 75 held-out primary-domain validation episodes.
\end{samepage}

More demonstrations improve grasping without resolving return timing. Under matched optimization budgets, scaling $\pi_{0.5}$ from 758 to 1,453 demonstrations raises grasp success from 65.2\% to 73.3\%, but post-grasp success changes only from 33.3\% to 36.4\%, and MacroMAE remains high at 72.9~g.

With prescribed motion in original layouts, TR-LT calibration reduces MacroMAE from 95.56 to 19.87~g. \methodname{} has lower error at every target while learning manipulation from randomized source positions. The remaining TR-LT error indicates that wrist-load change alone is insufficient to estimate transferred mass during dynamic pouring: container orientation, arm motion, vessel oscillation, and liquid sloshing all affect the measured wrench.

The ablations distinguish errors in action completion from errors in quantity regulation (Table~\ref{tab:ablations}). Grasp success remains comparable across variants (65.6--69.8\%, versus 68.3\% for the full model). Removing boundary-aware supervision raises MacroMAE from 13.15 to 20.27~g and reduces post-grasp success from 94.0\% to 84.6\%. Partial return followed by renewed pouring under the POUR prompt occurred in 9 of the 33 completed trials, with multiple reversals in 4.

The gate and trigger variants retain high post-grasp success (94.6--97.5\%), but their mass errors increase. The gate addresses integration drift: without it, MacroMAE reaches 16.92~g and bias shifts from $+2.65$ to $-5.73$~g, consistent with false accumulated mass prompting early return.

Tail prediction addresses delayed transfer. With policy and mass estimator fixed, current-mass, fixed $(0.2\,\mathrm{s})\widehat f_t$, and learned-tail triggers yield 19.06, 15.79, and 13.15~g MacroMAE. In Fig.~\ref{fig:closed_loop_300g}, predicted residual mass triggers return before estimated mass reaches 300~g. The final mass is 298.60~g.

\FloatBarrier
\subsection{Held-Out Goals and Additional Domains}

\begin{figure}[t]
    \centering
    \includegraphics[width=\columnwidth]{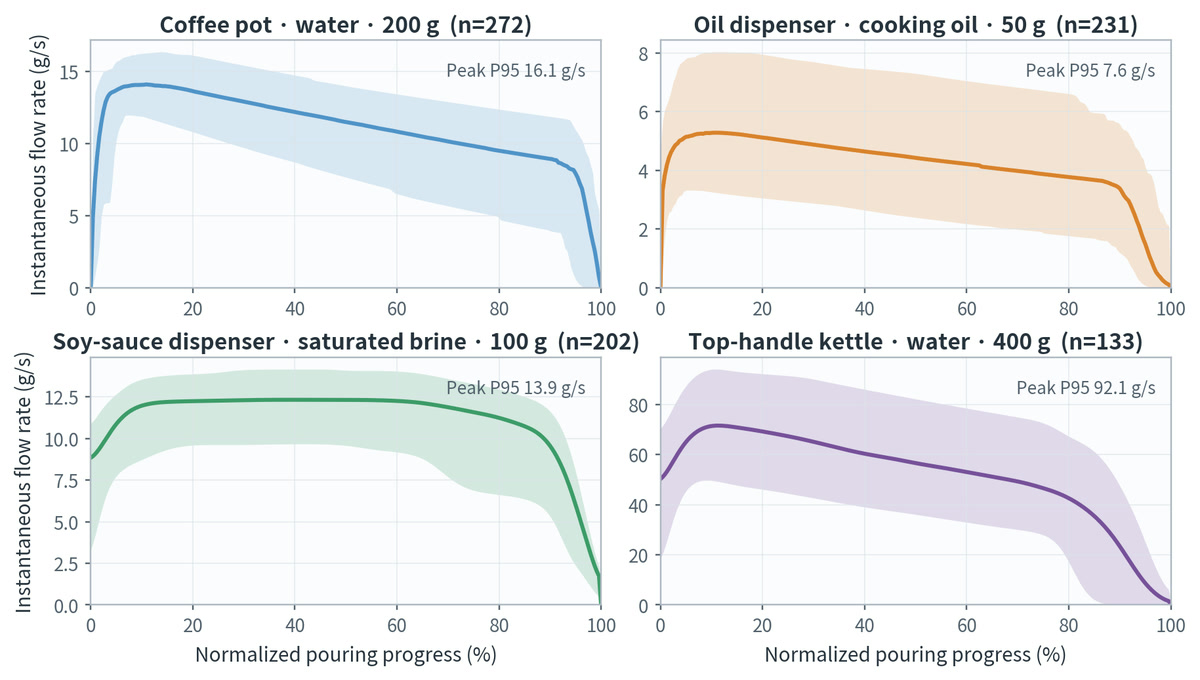}
    \caption{Median flow (lines) and central 95\% episode ranges (shading), with time normalized from onset (0\%) to cutoff (100\%). Peak P95 is the 95th percentile of episode peak flow.}
    \label{fig:task_flow_profiles}
\end{figure}

Target-independent estimation supports numeric interpolation. At 150/250~g, changing only the primary model's POUR prompt and completion threshold gives 7.6\% goal-balanced relative MAE and 54.8\% overall quantitative success (Table~\ref{tab:domain_results}).

Each additional task uses a separately trained domain-specific model to evaluate the completion formulation across distinct flow regimes. Figures~\ref{fig:training_flow_distribution} and~\ref{fig:task_flow_profiles} show fast kettle flow and slower cooking-oil flow; cooking oil also has a longer return tail. Among the additional tasks, the top-handle kettle has the highest grasp success (78.1\%) and E2E success (75.0\%). Its wide opening produces the fastest observed flow profile, and its 15.0\% goal-balanced relative MAE is also the highest of the three, consistent with greater sensitivity to return timing. The oil task reaches 57.4\% E2E success.

\pagebreak[3]
Saturated brine yields 9.05~g completed-trial MAE but 35.5\% E2E success. Failures mainly involve difficult grasping and spilling from the soy-sauce dispenser's spout.

\vfill
\pagebreak[4]
\section{Conclusion}
\label{sec:conclusion}

Physical-completion aliasing means nearly unchanged images and robot poses can require different actions as mass accumulates. Our experiments show that more demonstrations and force feedback do not by themselves yield precise quantitative stopping. \methodname{} couples boundary-aware action learning with persistent mass estimation and residual-outflow prediction, triggering return based on predicted terminal mass. These results support modeling completion as a persistent physical state, separately from motor generation. Target-independent estimation enables held-out target interpolation, while independently trained models cover distinct flow regimes. The approach still requires domain-specific training and offline electronic-scale supervision. Future work should learn transferable flow, mass, and residual-outflow representations to reduce retraining and dependence on instrumented demonstrations.

\FloatBarrier

\bibliographystyle{IEEEtran}
\bibliography{references}

\begin{thebibliography}{10}
\providecommand{\url}[1]{#1}
\csname url@samestyle\endcsname
\providecommand{\newblock}{\relax}
\providecommand{\bibinfo}[2]{#2}
\providecommand{\BIBentrySTDinterwordspacing}{\spaceskip=0pt\relax}
\providecommand{\BIBentryALTinterwordstretchfactor}{4}
\providecommand{\BIBentryALTinterwordspacing}{\spaceskip=\fontdimen2\font plus
\BIBentryALTinterwordstretchfactor\fontdimen3\font minus
  \fontdimen4\font\relax}
\providecommand{\BIBforeignlanguage}[2]{{%
\expandafter\ifx\csname l@#1\endcsname\relax
\typeout{** WARNING: IEEEtran.bst: No hyphenation pattern has been}%
\typeout{** loaded for the language `#1'. Using the pattern for}%
\typeout{** the default language instead.}%
\else
\language=\csname l@#1\endcsname
\fi
#2}}
\providecommand{\BIBdecl}{\relax}
\BIBdecl

\bibitem{zitkovich2023rt2}
\BIBentryALTinterwordspacing
B.~Zitkovich, T.~Yu, S.~Xu \emph{et~al.}, ``{RT-2}: Vision-language-action
  models transfer web knowledge to robotic control,'' in \emph{Proceedings of
  the 7th Conference on Robot Learning}, ser. Proceedings of Machine Learning
  Research, vol. 229, 2023, pp. 2165--2183. [Online]. Available:
  \url{https://proceedings.mlr.press/v229/zitkovich23a.html}
\BIBentrySTDinterwordspacing

\bibitem{black2025pi05}
{Physical Intelligence}, K.~Black, N.~Brown \emph{et~al.}, ``$\pi_{0.5}$: A
  vision-language-action model with open-world generalization,''
  arXiv:2504.16054, 2025.

\bibitem{bjorck2025groot}
{NVIDIA}, J.~Bjorck, F.~Casta{\~n}eda \emph{et~al.}, ``{GR00T N1}: An open
  foundation model for generalist humanoid robots,'' arXiv:2503.14734, 2025.

\bibitem{piacenza2022pouring}
P.~Piacenza, D.~Lee, and V.~Isler, ``Pouring by feel: An analysis of tactile
  and proprioceptive sensing for accurate pouring,'' in \emph{2022
  International Conference on Robotics and Automation (ICRA)}, 2022, pp.
  10\,248--10\,254.

\bibitem{zhu2023pour}
F.~Zhu, S.~Hu, L.~Leng, A.~Bartsch, A.~George, and A.~Barati~Farimani, ``Pour
  me a drink: Robotic precision pouring carbonated beverages into transparent
  containers,'' arXiv:2309.08892, 2023.

\bibitem{yu2025forcevla}
J.~Yu, H.~Liu, Q.~Yu, J.~Ren, C.~Hao, H.~Ding, G.~Huang, G.~Huang, Y.~Song,
  P.~Cai, W.~Zhang, and C.~Lu, ``{ForceVLA}: Enhancing vla models with a
  force-aware {MoE} for contact-rich manipulation,'' in \emph{Advances in
  Neural Information Processing Systems}, vol.~38, 2025.

\bibitem{zhang2025tavla}
Z.~Zhang, H.~Xu, Z.~Yang, C.~Yue, Z.~Lin, H.-a. Gao, Z.~Wang, and H.~Zhao,
  ``{TA-VLA}: Elucidating the design space of torque-aware
  vision-language-action models,'' in \emph{Proceedings of the 9th Conference
  on Robot Learning (CoRL)}, 2025.

\bibitem{xue2025rdp}
H.~Xue, J.~Ren, W.~Chen, G.~Zhang, Y.~Fang, G.~Gu, H.~Xu, and C.~Lu, ``Reactive
  diffusion policy: Slow-fast visual-tactile policy learning for contact-rich
  manipulation,'' in \emph{Proceedings of Robotics: Science and Systems}, 2025.

\bibitem{li2026fmvla}
R.~Li, Q.~Li, R.~Ma, Y.~Deng, L.~Luo, Z.~Du, J.~Xiang, H.~Liang, R.~Wang,
  J.~Yang, and B.~Guo, ``{FM-VLA}: Force-based memory for
  vision-language-action models in contact-rich manipulation,''
  arXiv:2607.18231, 2026.

\bibitem{schenck2017visual}
C.~Schenck and D.~Fox, ``Visual closed-loop control for pouring liquids,'' in
  \emph{2017 IEEE International Conference on Robotics and Automation (ICRA)},
  2017, pp. 2629--2636.

\bibitem{hu2025robocap}
Y.~Hu, A.~Gillespie, A.~Padmanabha, K.~Puthuveetil, W.~Lewis, K.~Khokar, and
  Z.~Erickson, ``{RoboCAP}: Robotic classification and precision pouring of
  diverse liquids and granular media with capacitive sensing,'' in \emph{2025
  IEEE/RSJ International Conference on Intelligent Robots and Systems (IROS)},
  2025, pp. 19\,546--19\,553.

\bibitem{kennedy2019autonomous}
M.~Kennedy, K.~Schmeckpeper, D.~Thakur, C.~Jiang, V.~Kumar, and K.~Daniilidis,
  ``Autonomous precision pouring from unknown containers,'' \emph{IEEE Robotics
  and Automation Letters}, vol.~4, no.~3, pp. 2317--2324, 2019.

\bibitem{pan2016motion}
Z.~Pan and D.~Manocha, ``Motion planning for fluid manipulation using
  simplified dynamics,'' in \emph{2016 IEEE/RSJ International Conference on
  Intelligent Robots and Systems (IROS)}, 2016, pp. 4224--4231.

\bibitem{huang2021pouring}
Y.~Huang, J.~Wilches, and Y.~Sun, ``Robot gaining accurate pouring skills
  through self-supervised learning and generalization,'' \emph{Robotics and
  Autonomous Systems}, vol. 136, p. 103692, 2021.

\bibitem{noda2007pouring}
Y.~Noda and K.~Terashima, ``Modeling and feedforward flow rate control of
  automatic pouring system with real ladle,'' \emph{Journal of Robotics and
  Mechatronics}, vol.~19, no.~2, pp. 205--211, 2007.

\bibitem{sueki2021flow}
Y.~Sueki and Y.~Noda, ``Experimental verification of real-time flow-rate
  estimations in a tilting-ladle-type automatic pouring machine,''
  \emph{Applied Sciences}, vol.~11, no.~15, p. 6701, 2021.

\bibitem{chi2023diffusionpolicy}
\BIBentryALTinterwordspacing
C.~Chi, S.~Feng, Y.~Du, Z.~Xu, E.~Cousineau, B.~Burchfiel, and S.~Song,
  ``Diffusion policy: Visuomotor policy learning via action diffusion,'' in
  \emph{Proceedings of Robotics: Science and Systems}, 2023. [Online].
  Available: \url{https://diffusion-policy.cs.columbia.edu/}
\BIBentrySTDinterwordspacing

\bibitem{li2026forcevla2}
Y.~Li, Z.~Zhaxizhuoma, H.~Jiang, J.~Xia, H.~Zhang, J.~Du, Y.~Zhou, J.~Zeng,
  C.~Hao, J.~Ren, Q.~Yu, C.~Lu, Y.~Qiao, and J.~Pang, ``{ForceVLA2}: Unleashing
  hybrid force-position control with force awareness for contact-rich
  manipulation,'' in \emph{Proceedings of the IEEE/CVF Conference on Computer
  Vision and Pattern Recognition (CVPR)}, 2026, pp. 8911--8920.

\bibitem{kim2026progvla}
S.~Kim, J.~Choi, S.~Baek, and J.-M. Renders, ``{ProgVLA}: Progress-aware robot
  manipulation skill learning,'' arXiv:2605.28231, 2026.

\bibitem{yan2026progressvla}
H.~Yan, Q.~Li, J.~Yang, and Y.~Mu, ``{ProgressVLA}: Progress-guided diffusion
  policy for vision-language robotic manipulation,'' arXiv:2603.27670, 2026.

\bibitem{lee2020multimodal}
M.~A. Lee, B.~Yi, R.~Mart{\'i}n-Mart{\'i}n, S.~Savarese, and J.~Bohg,
  ``Multimodal sensor fusion with differentiable filters,'' in \emph{2020
  IEEE/RSJ International Conference on Intelligent Robots and Systems (IROS)},
  2020.

\bibitem{revach2022kalmannet}
G.~Revach, N.~Shlezinger, X.~Ni, A.~L. Escoriza, R.~J.~G. van Sloun, and Y.~C.
  Eldar, ``{KalmanNet}: Neural network aided kalman filtering for partially
  known dynamics,'' \emph{IEEE Transactions on Signal Processing}, vol.~70, pp.
  1532--1547, 2022.

\bibitem{cho2014learning}
K.~Cho, B.~van Merri{\"e}nboer, C.~Gulcehre, D.~Bahdanau, F.~Bougares,
  H.~Schwenk, and Y.~Bengio, ``Learning phrase representations using {RNN}
  encoder--decoder for statistical machine translation,'' in \emph{Proc.
  EMNLP}, 2014, pp. 1724--1734.

\end{thebibliography}

\end{document}